# ARCO1: An Application of Belief Networks to the Oil Market

Bruce Abramson
University of Southern California
Department of Computer Science
Los Angeles, CA 90089-0782

## Abstract

Belief networks are a new, potentially important, class of knowledge-based models. ARCO1, currently under development at the Atlantic Richfield Company (ARCO) and the University of Southern California (USC), is the most advanced reported implementation of these models in a financial forecasting setting. ARCO1's underlying belief network models the variables believed to have an impact on the crude oil market. A pictorial market model—developed on a MAC II—facilitates consensus among the members of the forecasting team. The system forecasts crude oil prices via Monte Carlo analyses of the network. Several different models of the oil market have been developed; the system's ability to be updated quickly highlights its flexibility.

## 1 Introduction

Belief networks are a class of models that have recently become important to researchers at the intersection of artificial intelligence (AI) and decision analysis (DA). Despite their underlying sophistication, belief networks are conceptually simple. Any directed acyclic graph (DAG) in which (i) nodes represent individual variables, items, characteristics, or knowledge sources, (ii) arcs demonstrate influence among the nodes, and (iii) functions associated with the arcs indicate the nature of that influence, qualifies as a belief network (Abramson 1990). Belief networks were originally introduced as a middle ground between psychologically valid elicitation procedures and mathematically valid representations of uncertainty (Howard and Matheson 1984). As such, they begin with an understanding of the heuristics and biases that typically plague experts (Kahneman, Slovic, and Tversky 1982), the DA elicitation techniques that help overcome these biases (von Winterfeldt and Edwards 1986), and the axioms of Bayesian probability theory (Savage 1954, Edwards, Lindman, and Savage 1963). These basic principles have led to a variety of inference and decision algorithms (Pearl 1988, Shachter 1986, 1988).

Several powerful belief network-based systems have been discussed in the literature. The two most developed of these systems, MUNIN (Andreassen et. al. 1987) and Pathfinder (Heckerman, Horvitz, and Nathwani 1990), deal with medical diagnoses. ARCO1 marks the first reported forecasting application. This paper outlines the modeling effort that went into ARCO1, and reports its preliminary models of the 1990 oil market and its forecasts of 1990 prices. For a more detailed treatment of this material, see (Abramson and Finizza in press).

## 2 Domain Specifics

Models of the world oil market can be broadly classified into as either optimization models or target capacity utilization (TCU) models (Energy Modeling Forum 1982, Gately 1984, Powell 1990). Optimization models, which are generally based on economic theories of depletable resources and/or cartels, are used primarily for long term projections (Marshalla and Nesbitt, 1986). Since our aim was to develop a system for short term forecasts, we chose to develop ARCO1 as a TCU model.

The central determinant of prices in a TCU framework is the relationship of calculated production to exogenously determined capacity; the resulting measure of market tightness indicates price pressure. Our choice of this framework as the basis of ARCO1's model stresses the importance of subjective political variables. Since capacity is exogenously determined and short term crude oil demand is highly price inelastic and almost completely specified by seasonal patterns, short term price forecasts can be (more or less) reduced to forecasts of production. Production levels, in turn, are essentially set by the political decisions of the governments of oil producing countries;

OPEC's Persian Gulf members (Saudi Arabia, Iran, Iraq, Kuwait, UAE, and Qatar) are particularly important, because they tend to be the only producers with substantial slack capacity. Thus, the inclusion of political analyses and adjustments to *production* calculations appear to be much more appropriate than politically motivated judgemental adjustments to mechanically forecast *price* calculations.

ARCO1's base case was a model of the 1990 oil market designed in early 1990 using historical data through the fourth quarter of 1989 and subjective assessments provided between November 1989 and February 1990. This model, depicted in Figure 1, has already undergone revisions and will continue to be revised. It contains about 140 equations, many with time lags and some expressed as conditional probabilities. Section 3 enumerates these variables and relationships. It is important to stress, however, that the system is more than simply a model. ARCO1 was designed to facilitate scenario development and simulation exercises. One such exercise is discussed in Section 4; it concerns a scenario developed in late-August/early-September 1990 to reflect the altered political realities of the Persian Gulf.

## 3  Model Variables

This section explains ARCO1's variables, as shown in Figures 1 and 2. The variables can be broken into seven categories and eight time periods. The time periods range, by quarter, from the first quarter of 1989 through the fourth quarter of 1990. The categories, in turn, are historical, annual, tax, demand, supply, politics, and price.

**Historical Variables**   represent events that have already occurred; their values were retrieved from the appropriate references.

**Annual Variables**   are not expected to change over the course of the year.

> **NC Cap:** physical production capacity of non-core OPEC countries (i.e., OPEC countries outside the Persian Gulf).
> **NC Prod:** actual production of non-core OPEC countries.
> **World Growth:** world GDP growth is broken into four components: lesser developed countries (LDC), Western Europe (WE), US, and Japan. Coefficients relating the four components of the world economy to the single world growth variable were calculated by linear regression.

**Tax Variables**   relate to US tax policy. Two types of potentially relevant legislation are envisioned: an oil import fee (OI Fee) and an increase in the federal gasoline tax (GT indicates whether or not the tax will be passed. GT Impact translates the tax from dollars-per-gallon to dollars-per-barrel). If an oil import fee is imposed, its presumed effect would be to place an $18 per barrel floor on imported oil prices. Increases in the federal gasoline tax could range from $.01 to $.50 per gallon.

**Demand Variables**   are used to calculate total free world demand. Demand calculations, (at least in the developed world), are more straightforward than supply calculations because there are fewer phenomena that allow a small group of decision makers to affect the market. The only demand-side peculiarity identified, in fact, was fuel switching, a decision on the part of the managers of dual-fired utility plants to switch to oil use; its impact is restricted to times when prices are maintained below $15 per barrel, and may range as high as 2 MMBD (million barrels per day).

> **Level:** prevailing price at start of quarter.
> **Duration:** length of time over which the current price level has prevailed.
> **Fuel Switching:** amount of increased demand due to the adoption of oil by utility plants with dual-fired furnaces. Conditionally dependent on price level and duration.
> **Demand:** total world demand, by quarter. Functional specification was determined by an *ad hoc* combination of regression techniques and scenario analysis.

**Supply Variables**   are used to calculate free world supply.

> **US Prod:** US production levels.
> **NO Prod:** other non-OPEC (non-US) production.
> **C Cap:** physical production capacity of core OPEC countries.
> **Delta I:** the change in inventory levels.
> **O Call:** effective demand for OPEC oil, or the call on OPEC. Defined as total world demand minus oil supplied by other sources.
> **Core Demand:** effective demand for oil from core OPEC countries, defined as the amount demanded from OPEC minus the amount supplied by non-core countries.
> **Core Production:** amount produced by the core OPEC countries, given as core demand plus a "political hedge factor."
> **Cap Ut:** defined as the percentage of core OPEC capacity being used for production.
> **Supply:** total world supply of crude.
> **DeltaY Core Prod:** one year change in production by core OPEC countries.

**DeltaQ Core Prod:** one quarter change in production by core OPEC countries.

**DeltaY Sweet:** the one year change in production of light, sweet crude, (all non-OPEC production).

**Political Variables** were introduced to capture subjective measures of core OPEC politics. Two different aspects of the political situation are considered: general intra-gulf relations (Intragulf) and the degree of conflict arising from unhappiness with market share (Market Share). These two variables are then combined and mapped into production to yield a "political hedge to production" factor (Politics).

**Price Variables** directly represent the price of various grades of crude, OPEC (Saudi basket) and WTI (West Texas Intermediate, the benchmark US crude).

**OPEC:** the price of OPEC oil, given as a regression-weighted function.

**Time:** introduced to measure the trend of a steadily increasing shortage of sweet crude.

**SS Diff:** difference in price between sweet and sour crude. It is measured by a complex formula.

**WTI:** price of WTI oil, subject to the possible imposition of an oil import fee.

## 4  Scenarios

The previous section defined the variables and influences captured by ARCO1's model of the 1990 oil market. Value ranges and precise dependence (algebraic, econometric, and probabilistic) were omitted from the discussion, as they are from Figure 1's picture of the network; they are neither central to the model of the domain nor of particular interest to most AI researchers. They are, however, crucial if the model is to produce any useful results. This distinction is characteristic of belief networks; network structure (i.e., nodes and arcs) describes the domain, while network parameters (i.e., historical data, prior probabilities, and numeric relationships) allow specific questions to be answered. Viewed another way, the model illustrated in Figure 1 captures one year of the oil market. A fixed set of parameters (such as those that we used in our studies) captures a time frame (the year 1990). Thus, recasting the model for 1991 should require nothing more than reviewing and updating the network's parameters. (It would, in fact, be this simple were the network structure completely satisfactory. Several potentially hazy areas—notably US tax policy, inventory behavior, and Gulf politics—have already been detected, and are currently under revision. Once a fully satisfactory network structure has been derived, however, updates should be restricted to parameter changes. Structural changes should be few and far between, and should correspond to fundamental changes in the market).

The discussion of variables and influence, then, was intended to convey a broad understanding of the oil market. Forecasts, on the other hand, require data. The basic model was used to create two sets of scenarios for 1990: a base case and a constrained capacity case. The base case was designed in early 1990, and covers all four quarters of the year. The constrained capacity case was designed in late August/early September 1990, when a fundamental market shift occurred; it assumes an effective boycott of Iraqi and Kuwaiti oil, and that all other producers produce at maximum capacity. A revised network, shown in Figure 2, was constructed. It accounted for historical data through the end of the second quarter, and produced forecasts for the third and fourth quarters of 1990. These cases were developed to demonstrate the system's flexibility, not its accuracy. Recall that the system's processing power is still restricted to Monte Carlo analyses; this entire phase of development must be viewed more as a proof-of-concept than as a demonstration of power.

### 4.1  Base Case

The base case for 1990 is described by the network of Figure 1. Specific values for historical and exogenous variables were retrieved from the appropriate sources. For further details and the actual values assigned to these variables, see (Abramson and Finizza in press). Many of the probabilistic assessments and regression weights are currently under review and have yet to be released. Qualitative analyses of US tax policy, fuel switching, and Gulf politics, however, certainly warrant further discussion.

US tax policy is one instance of an important set of judgemental variables that doesn't fit into data-driven models. Since the US is the world's largest consumer of oil, as well as its largest importer and one of its largest producers, US policies can affect the market in several ways. First, increased taxation could lead to slowly declining demand. Second, US taxes could have a direct impact on the price of imported oil, consequently an indirect impact on world prices and on domestically produced crude. Since crude oil spot prices are typically quoted for WTI, (as traded on NYMEX, the New York Mercantile Exchange), the price of domestic oil is of central importance. The analysis of US tax policy considered the possibility of two relevant taxes: an increase in the federal gasoline tax, and the imposition of an oil import fee (at an $18 floor). The general assessment was that the increased gasoline tax was the more likely of the two, and that if passed, would decrease the likelihood of an oil import fee. (Note that the federal gasoline tax was, in fact, increased for 1991 as part of the budget package eventually passed in October 1990).

Fuel switching, (a managerial decision at dual-fired utility plants to burn oil rather than natural gas), is another area that generally eludes data-driven models. This decision, unlike most others affecting the market, is an essentially macro-level demand decision; universal switching to oil could increase demand by as much as 2 MMBD. ARCO1's model included a fairly detailed analysis of fuel switching. Without getting into specifics, the thrust of the analysis is that oil warrants consideration at prices below $15 per barrel. If low prices are maintained for an extended period of time, many managers will opt for oil. The lower the prices and the longer they are maintained, the greater the demand.

Political analyses and projections are necessarily softer (i.e., more subjective) than their economic counterparts. As a result, they are invariably omitted from technical models, and relegated to the role of *a posteriori* judgemental adjustments. ARCO1's structure helped initiate a quantitative—albeit subjective—analysis of core OPEC (Persian Gulf) politics. The first step in the analysis lay in realizing that politics is really significant only as it affects production. Thus, rather than being an adjustment to price, politics is viewed as an adjustment to OPEC production. The second step used this observation to identify two relevant characteristics of Gulf politics: (i) general political amicability among the core OPEC members, and (ii) each country's satisfaction with its market share (and thus, implicitly, compliance with OPEC quotas). The third step placed these variables on a subjective five-point scales (harmony-to-war and strict compliance-to-rampant cheating, respectively). The fourth step specified conditional probabilities relating the two, and the fifth step mapped them into oil production above or below natural demand. This type of political analysis is inexact, and will certainly need to be refined, updated, and changed. (One such change is discussed in the next section). Its inclusion in the model, however, establishes a clear "political module" into which all updates can easily be inserted.

### 4.2   Constrained Capacity Case

The constrained capacity (or boycott) case was designed after Iraq's invasion of Kuwait on August 2, 1990. The behavior of the oil market following the Iraqi invasion and the subsequent world reaction indicated that a fundamental change had occurred, and that all existing short-term models were in need of (at least some) revision. The modularity of ARCO1's underlying network facilitated these changes; the constrained capacity network is shown in Figure 2.

The network that corresponds to this case incorporated several assumptions that actually made the analysis *easier* than it was in the base case. First, (and most obviously), it was designed seven months later and with two additional quarters of historical data.

Thus, nodes corresponding to first and second quarter 1989 were dropped, and actual numbers for first and second quarters 1990 were included. Second, it began with the assumptions that the boycott of Iraq and Kuwait would be effective, and that everyone else in the world would raise their production levels to their maximum physical capacity. Since all non-OPEC producers were already assumed to be producing to capacity, and the production levels of non-core OPEC countries had been fixed at 90% of their capacity, few changes were needed outside OPEC's core. Core production, however, was pushed up to the combined capacities of Saudi Arabia, Iran, UAE, and Qatar. Capacity utilization, originally introduced as a measure of pressure on production and the key to the model, was fixed at 1.0. As a result, the entire political analysis module was dropped from the network; despite the obvious volatility of the political situation, the impact on production was assumed to be steady throughout the rest of the year. (The model assumed that no settlement negotiated before the end of the year would restore the situation *ante*). Fuel switching was dropped from the analysis because the possibility of sustained prices under $15 per barrel disappeared, and US tax policy was excluded (perhaps unreasonably) because it did not appear likely to have much of an impact before the end of 1990. Most of the other analyses remained as they had been in the base case.

One point worth noting is that this constrained capacity case is significantly outside the range of possibilities that were envisioned when the base case was designed. The initial political assessment, in fact, assumed that the pattern set during the Iran/Iraq war would continue: lack of cohesion among Gulf countries would lead to overproduction and low prices. The possibility of a consumers boycott was not even considered. Nevertheless, the model was flexible enough to be updated (quickly and painlessly) in the presence of new data.

## 5   Forecasts

The first phase of ARCO1's development stressed model construction. The system's ultimate objective, however, is to *use* these models to forecast the market. ARCO1's underlying belief networks captured information about direct interrelationships among the variables affecting the oil market. Implicit in these direct relationships lies information about the market's indirect relationships. The task of the forecasting/processing engine must be to explicate the indirect relationships between exogenous variables and future prices.

Although a wide range of statistical procedures are (theoretically) available to ARCO1, only one simple technique has been fully implemented to date: Monte Carlo analysis. The implementation of Monte Carlo on the network was fairly straightforward. Exoge-nous variables, (represented by rooted nodes, or nodes with no ingoing arcs), were specified as either constant values or as unconditional (prior) probability distributions. In either case, assigning a single value to an exogenous variable was straightforward. Once all rooted nodes were instantiated, nodes pointed to only by rooted nodes, (i.e., variables directly dependent only on exogenous variables), could similarly be instantiated. This procedure continued until the entire network (or, alternatively, the mid-network node selected as the forecast's target) was instantiated. This assignment of a single value to each variable constitutes a single fully-specified scenario (i.e., all variables are instantiated); the procedure is guaranteed to terminate because a belief network is a DAG (i.e., it contains no cycles). Multiple fully-specified scenarios lead to a distribution of values across the target variable, and thus a probabilistically reported forecast.

Results were generated by Monte Carlo analyses of the 1990 networks; they are not meant to be either complete or conclusive, but simply illustrative of the claim that the system works. The variables targeted by these forecasts were the network's sinks, namely WTI or WTIp (by quarter). The WTIp variables recognize the possibility of an oil import fee placing an $18 floor on domestic oil prices; they adopt the price calculated for WTI if no fee is imposed, but report a price of $18 if the fee is passed and the calculated price is less than or equal to $18. Three sets of simulations were run. The first set studied the full base case, simulating the network shown in Figure 1. The second set retained the base case assumptions, but updated the network with actual data for the first two quarters of 1990. The third set simulated the constrained capacity case, using the network of Figure 2. (As the data will show, however, this third set of simulations was not really necessary). In each of these simulations, 100 scenarios were generated for each target variable. (Simulations of 100 scenarios are not really adequate. The small size was necessitated by implementation inefficiencies. Many of them have already been corrected; our current implementation is running ten to twenty times as quickly). The results of these simulations are shown in Table 1.

The 1990 base case forecast indicated a relatively flat market. All four quarters generated average prices between $20 and $22, with an annual average of $21.14. The [$18,$21] range accounted for 246 of the 400 scenarios (61.5%), with just over half of them (50.5%) falling between $19 and $21. The distribution of the remaining 154 scenarios, however, was far from uniform. Only 34 scenarios (8.5%) projected prices at or below $17, and 23 of these actually hit the $17 level; the [$14,$16] range accounted for only 11 of 400 possible cases (2.75%). Thus, the probability of a significant downward trend under the base case conditions was highly unlikely. On the upside, however, there appeared to be more room for runaway prices. 67 scenarios (16.75%) generated prices in the [$22,$25] range, 43 in the [$26,$30] range (10.75%), and 10 in the [$31,$40] range (2.5%).

Recall that these results were based on assumptions available at the beginning of 1990. The actual average WTI price for the first quarter of 1990 was $21.70, within $1 (or about 1/3 of a standard deviation) from the forecast mean. In the second quarter, average WTI price was $17.76, about $3 (or one standard deviation) from the forecast mean. Thus, ARCO1's forecasts prior to the Iraqi invasion of Kuwait (and the ensuing fundamental shift in the market) were relatively accurate. The insertion of first and second quarter data, however, allowed us to re-run the simulations for the third and fourth quarter. These updates were produced using data available in July 1990. They are well within the range of projections made by most industry analysts at the time. Unfortunately, the market shifted sharply in August. When Iraq invaded Kuwait, the US successfully led the United Nations to establish an effective embargo of Iraqi and Kuwaiti oil, and all other producers decided to increase production, the global supply picture was altered drastically. Simulations of our constrained capacity case yielded the very tight forecasts shown in the final column of Table 1.

The conditions underlying the constrained capacity scenario are sufficiently restrictive to remove virtually all uncertainty from the system; detailed simulation and statistical analyses were unnecessary. Under its assumptions, supply is entirely fixed, and demand is assumed to vary more-or-less in line with world GDP growth. Thus, prices generated under this scenario (at least for the near term) are effectively fixed. Despite the volatility of spot prices throughout the third and fourth quarters of 1990, ARCO1's constrained capacity forecasts were remarkably accurate; the (true) average prices were $26.31 for the third quarter and $31.91 for the fourth. Technical volatility, however, does highlight a potential problem facing the system. The networks discussed in this paper all focus on market fundamentals. Volatility caused by war fears, unusually high risk factors, and other technical factors, tend to elude fundamental analyses. In a disequilibrated (or day-traded) market, forecasts produced by ARCO1 are unlikely to be useful. In a stable, fundamental-based setting, however, the information captured by ARCO1's network does appear to model our understanding of the crude oil market in a manner amenable to producing relatively accurate forecasts.

## 6 Conclusions

ARCO1 is a knowledge-based system designed to help the members of ARCO's corporate planning group who are involved with forecasting the price of crude oil. The system is based on a belief network, a type of graphical model that is rapidly gaining popularity in both the AI and DA research communities. ARCO1's construction was involved and time-consuming. As the first reported forecasting system of its type, it suggested many interesting basic research issues, most of which have yet to be explored. The underlying software is evolutionary; it grows in response to need. Since the first crucial stage of the system's development was the construction of a belief network model of the domain, work to date has emphasized modeling rather than forecasting. As a result, the forecasting applications may appear somewhat trivial—albeit surprisingly accurate. This paper was intended more as a proof-of-concept than as a demonstration-of-power. A great deal of evaluation—of both the underlying models and their forecasts—still need to be done.

| 1990 Base Case (original) | | 1990 Base Case (updated) | | Constrained Capacity Case |
|---|---|---|---|---|
| Quarter | $\mu$ | $\sigma$ | $\mu$ | $\sigma$ | Prices |
| 1Q 1990 | 20.87 | 2.9 | NA | NA | NA |
| 2Q 1990 | 20.62 | 3.3 | NA | NA | NA |
| 3Q 1990 | 21.23 | 4.1 | 19.18 | 2.5 | 25 |
| 4Q 1990 | 21.84 | 4.4 | 20.79 | 4.4 | 29-31 |

Table 1: Means and standard deviations of the forecasts generated by Monte Carlo analyses. All numbers in are approximate, and quoted in dollars per barrel.

## 7   Acknowledgements

Domain expertise was provided by ARCO's Anthony Finizza, Mikkal Herberg, Peter Jaquette, and Paul Tossetti. The code underlying the system was written by Keung-Chi Ng.

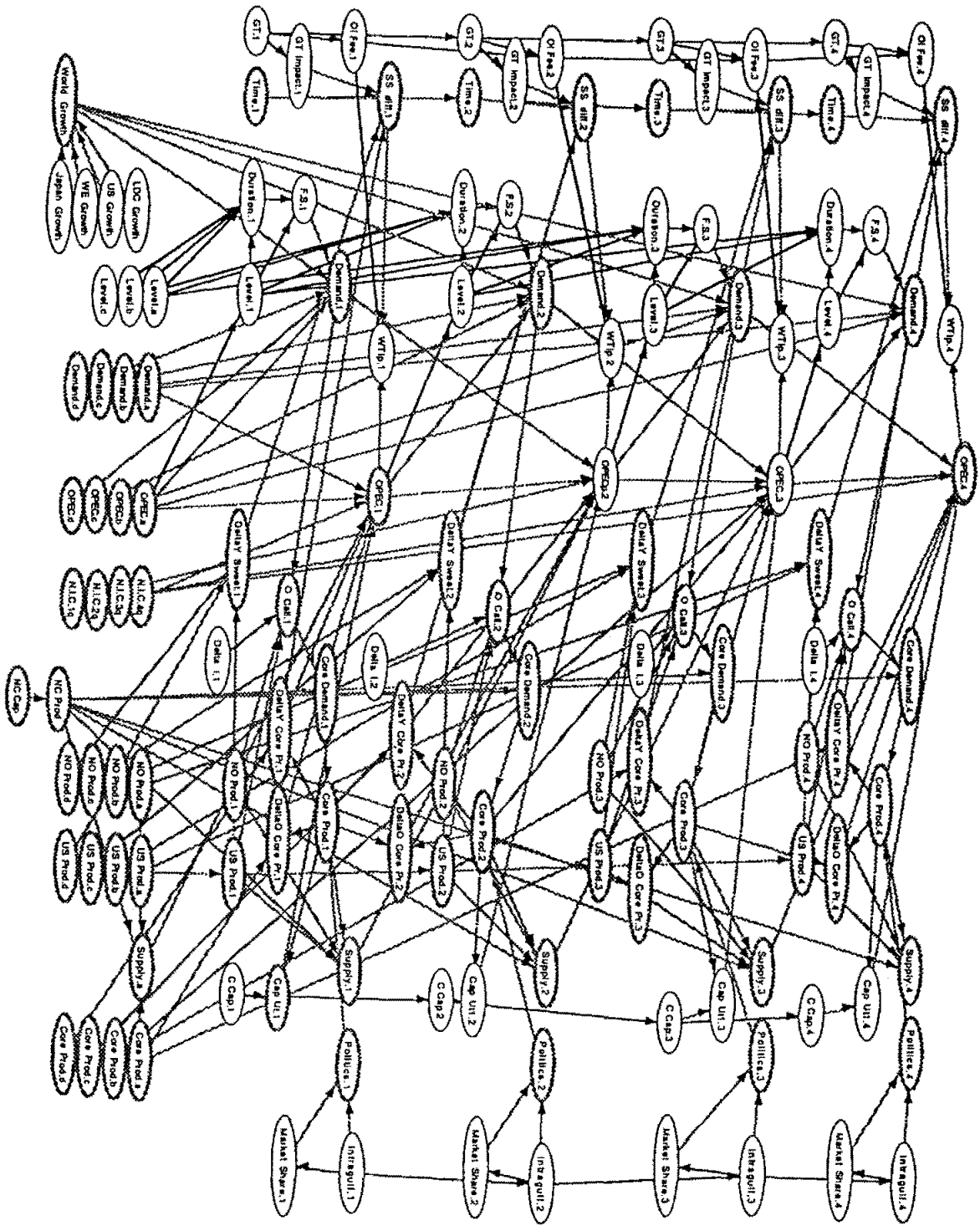

Figure 1: The network used to capture the 1990 base case. This model was designed in early 1990, using historical data through the fourth quarter of 1989 and subjective assessments provided by February 1990. Variables labelled ".1," ".2," ".3," and ".4" correspond to the first, second, third, and fourth quarters of 1990, respectively. Variables labelled ".d," ".c," ".b," and ".a" correspond to the first, second, third, and fourth quarters of 1989, respectively.

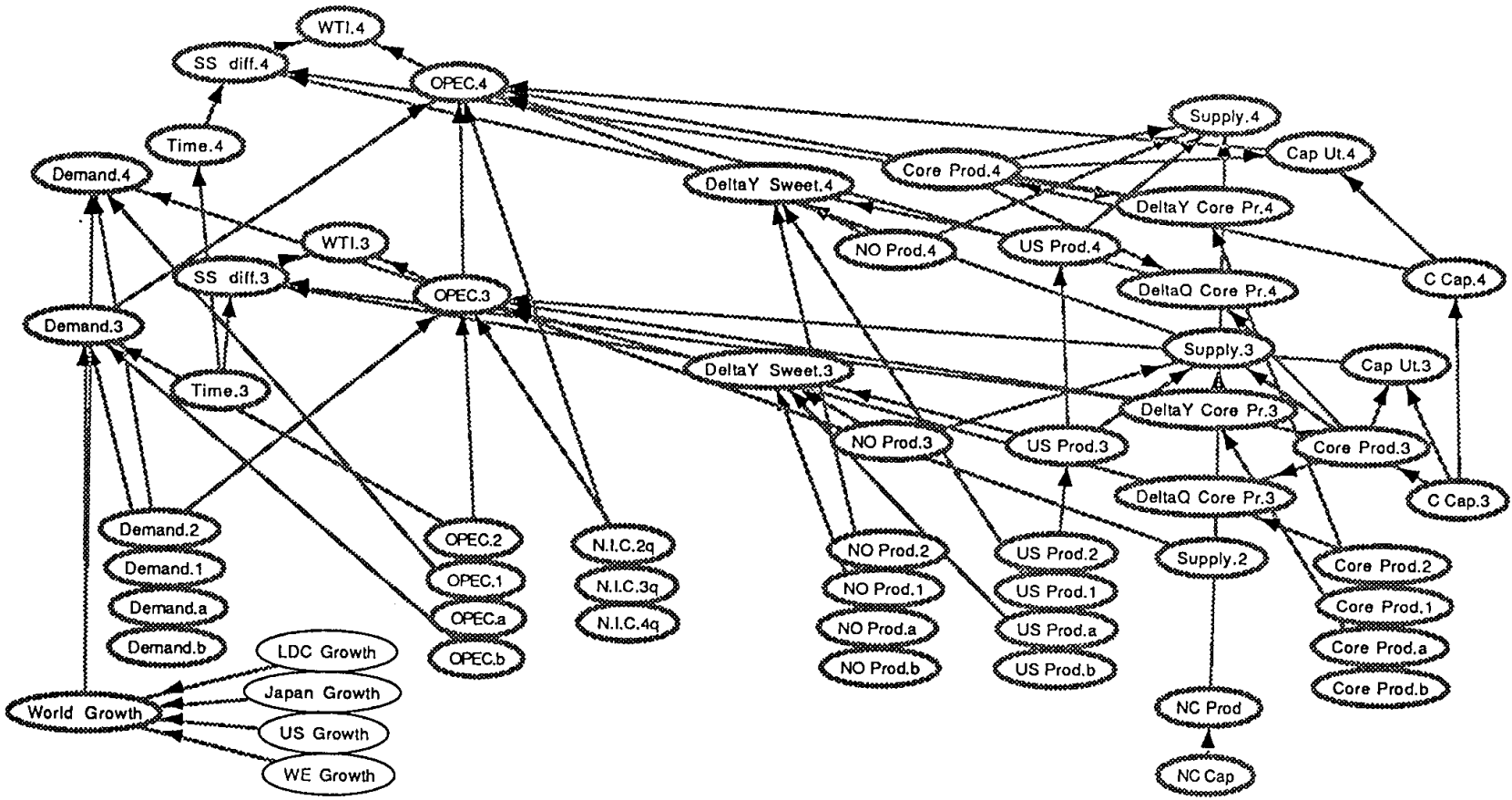

Figure 2: The network used to capture the constrained capacity case. This model was designed in September 1990, using historical data through the second quarter of 1989 and subjective assessments provided in August/September 1990. Variables labelled ".1," ".2," ".3," and ".4" correspond to the first, second, third, and fourth quarters of 1990, respectively. Variables labelled ".d," and ".c" correspond to the first, second, third, and fourth quarters of 1989, respectively.